%% file: root.tex
\documentclass[letterpaper, 10 pt, conference]{ieeeconf}  

\IEEEoverridecommandlockouts                              

\usepackage{graphics} 
\usepackage{epsfig} 
\usepackage{times} 
\usepackage{amsmath, bm} 
\usepackage{amssymb}  
\usepackage{dsfont}
\usepackage{makecell}
\usepackage{tabularx}
\usepackage{multirow}
\usepackage{booktabs}
\usepackage{censor}
\usepackage{cite}
\usepackage[normalem]{ulem}
\usepackage[colorlinks = true, linkcolor = black, urlcolor = black, citecolor = green, urlcolor = black]{hyperref}

\newcommand{\videourl}{\url{https://gianni0907.github.io/fault_tolerant_locomotion/}}
\graphicspath{{figures/}} 
\usepackage[nolist]{acronym} 
\input{tools/acronyms}
\input{tools/macros}

\title{\LARGE \bf
Learning Fault-Tolerant Locomotion with Adaptive Gait Timing
}

\author{Giovanbattista Gravina$^{1}$, Luca Rossini$^{1}$, Carlo Rizzardo$^{1}$, Arturo Laurenzi$^{1}$ and Nikos Tsagarakis$^{1}$%
\thanks{$^{1}$All authors are with the Humanoids and Human-Centered Mechatronics Research Line, Italian Institute of Technology, Genoa 16163, Italy (emails: {\tt\small name.surname@iit.it}). This work is supported by the EU Horizon 2020 EuROBIN project (grant agreement No.101070596).}%
}

\begin{document}

\maketitle
\thispagestyle{empty}
\pagestyle{empty}

\begin{abstract}

Hardware failures require legged robots to rapidly reorganize coordination and gait timing to maintain stability and mobility. This is particularly challenging for larger quadrupeds, where increased mass and tighter actuation limits reduce the feasibility of aggressive, high-frequency compensation strategies often observed on smaller platforms. In this work, we propose a deep reinforcement learning approach for fault-tolerant locomotion under actuator power loss. The method employs an asymmetric actor-critic architecture in which the critic has access to privileged information during training, while the actor learns to reconstruct a corresponding latent representation from proprioceptive observations. We introduce a latent-alignment loss that encourages consistency between actor and critic representations. Additionally, we augment the action space with a learnable gait frequency parameter, enabling adaptive gait timing in response to terrain variations and actuator degradation without predefined faulty-leg strategies. The approach is validated in high-fidelity simulation on uneven terrain and real-world experiments on flat ground using a 68\,kg quadruped robot.

\end{abstract}
\acresetall

\section{INTRODUCTION}
\label{sec:introduction}

Legged robots, especially quadrupeds \cite{hutter2016anymal, shin2022hound, kashiri2019centauro, valsecchi2023barry, rossini2026kyon}, are increasingly deployed in harsh and unstructured environments for real-world applications, such as search and rescue \cite{solmaz2024rescue}, inspection \cite{gehring2021inspection} and parcel delivery \cite{suarez2024eurobin, hooks2020alphred}. In such scenarios, the likelihood of hardware faults~---~like joint failures or actuator degradations~---~is significantly higher than in structured indoor or flat environments. Ensuring that a robot can continue to operate, or at least reach a safe state, while experiencing such degraded conditions is therefore a critical challenge for practical deployment.

This requires rapid adaptation to failures (e.g., transitioning from nominal to degraded gaits) and the capacity to encode a wide repertoire of behaviors within a single controller, as different faults may demand different strategies.
A robust controller must reorganize coordination across the remaining degrees of freedom, generate feasible motion patterns, and preserve stability, mobility, and task performance across diverse failure scenarios and terrains.

The mostly addressed hardware faults in the literature are: ($i$) joint locking fault \cite{yang2002fault}, where the joint becomes immobilized or suffers a severe reduction in its range of motion due to mechanical issues such as broken gears or transmission jamming, and ($ii$) power loss \cite{english1998fault,gor2018fault}, either partial, in case of weakened motor, or complete, when the actuator generates no torque causing the joint to rotate freely under external loads or gravity.

\begin{figure}[t]
    \includegraphics[width=\linewidth]{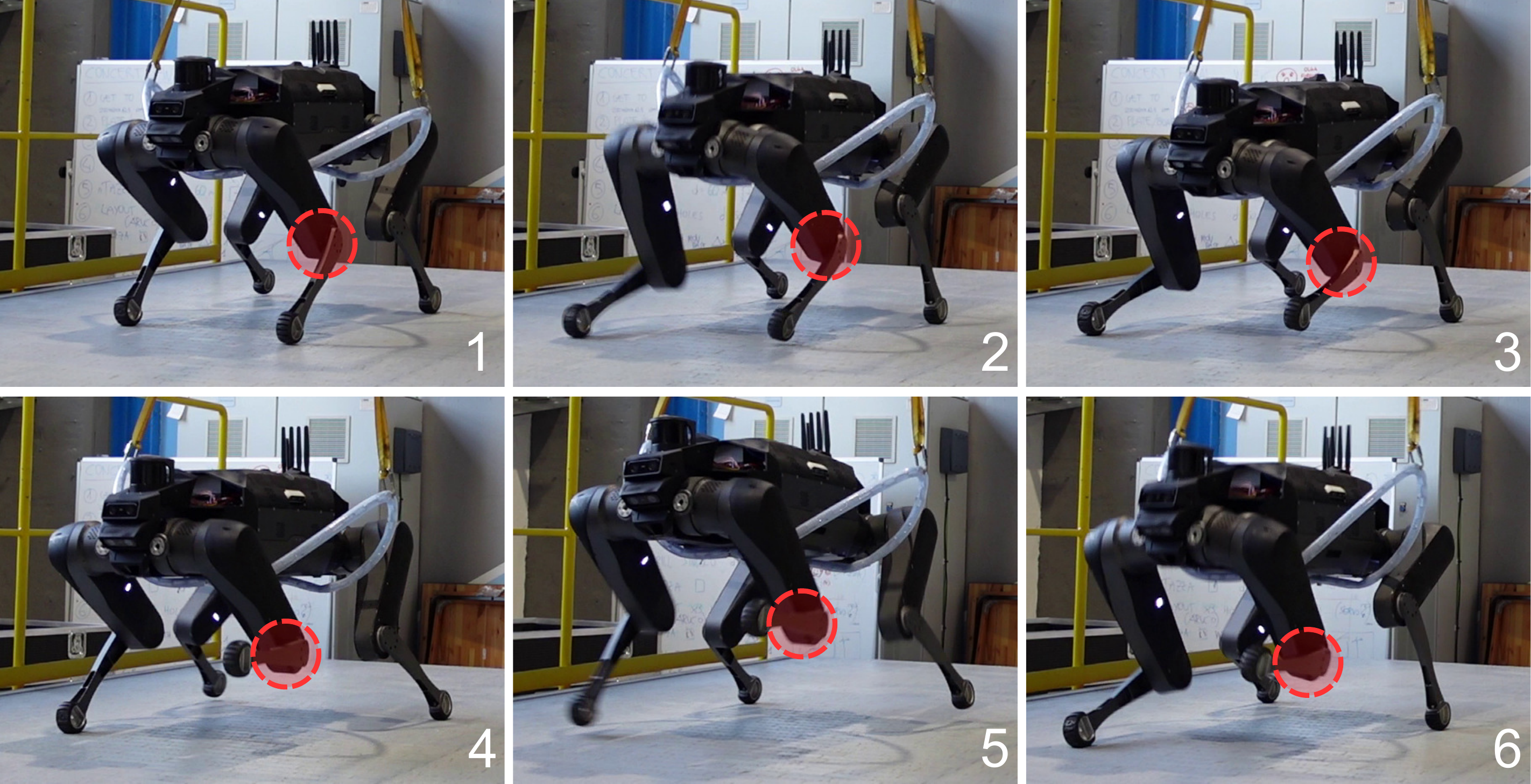}
    \caption{Fault-tolerant locomotion under sudden actuator power loss. The affected joint (highlighted in red) experiences a complete torque failure during motion. Experiments are performed on the 68\,kg Kyon quadruped robot \cite{rossini2026kyon}. Website: \videourl.}
    \label{fig:exp_flkp_snapshots}
\end{figure}


Recently, deep \ac{RL} has enabled control policies capable of addressing fault-tolerant locomotion by learning adaptive behaviors directly from data, reducing the reliance on manually designed, fault-specific control strategies.
Despite this progress, several open challenges remain. First, many prior approaches implicitly constrain locomotion under failure, either by relying on predefined gait generators for healthy legs or by discouraging the use of faulty legs. Such strategies may limit the policy's ability to fully exploit the remaining actuation capabilities left and to adapt the gait pattern to both fault conditions and terrain profile. Second, most experimental validations have been conducted on small-size quadrupeds and primarily on flat or mildly uneven terrain (e.g., grass, gravel or small height perturbations). Scaling these approaches to heavier legged robots~---~such as platforms designed for high payload and outdoor deployment~---~introduces additional challenges \cite{valsecchi2023barry}. 
Increased mass and inertia amplify dynamic coupling effects, reduce feasible stepping frequencies, and tighten actuation margins, making highly reactive or dynamically aggressive behaviors less viable, especially under hardware faults. 

In this work, we propose a framework to train a control policy via deep \ac{RL} to address fault-tolerant legged locomotion on uneven terrains under actuator power loss. The proposed architecture aims to reconstruct a latent representation of privileged observations, including joint fault status, by leveraging a history of proprioceptive observations. The main contributions of this paper are as follows:
\begin{itemize}
    \item Fault-aware locomotion from fault-unaware sensing: design of a single-stage training procedure based on an asymmetric actor--critic architecture, where the learning objective is augmented with an auxiliary latent-alignment loss that encourages the actor to reconstruct privileged representations from proprioceptive observations;
    
    \item Adaptive gait timing: augmentation of the action space with a learnable gait-frequency term alongside joint position targets, enabling the policy to autonomously regulate step timing and adapt contact scheduling to fault and terrain conditions;
    
    \item Fault inference: analysis of the effect of proprioceptive observation history length on the policy's ability to infer faults and compensate for them;
    
    \item Sim-to-real validation: evaluation of the proposed method in high-fidelity simulations on uneven terrain and real-world experiments on flat ground using the Kyon robot \cite{rossini2026kyon}, a 68\,kg medium-sized quadruped platform.
\end{itemize}

\section{RELATED WORK}
\label{sec:related}

\subsection{Reinforcement Learning for Legged Locomotion}
\label{subsec:loco}

Legged locomotion control has benefited significantly from recent developments in \ac{RL}. This approach has led to robust control policies for complex behaviors, including agile locomotion over uneven terrains \cite{lee2020learning, miki2022learning, choi2023learning}, high-speed locomotion \cite{margolis2024rapid}, and parkour skills \cite{cheng2024extreme, hoeller2024anymal}.

The possibility to avoid the high costs and physical risks associated with real-world experimentation by leveraging massively parallelized simulation environments to train agents safely is a major advantage of \ac{RL}. Training in simulation also provides access to privileged information which cannot be measured directly in real world scenarios and can be used to improve training efficiency and final performance. A common approach to utilize such privileged knowledge is through offline teacher-student distillation \cite{lee2020learning}. A teacher policy is first trained with access to all information, and then a student policy is trained to mimic the teacher using only sensors available on the robot. An alternative approach that retains the benefit of privileged information, but within a unified training loop, is the asymmetric actor-critic architecture \cite{pinto2017asymmetric}. In this case, the critic uses privileged information to evaluate actions during training, while the actor~---~which becomes the deployed controller~---~only uses real-world observations. 

While asymmetric actor-critic training is effective, a large gap between privileged and non-privileged observations can limit the actor’s ability to learn complex behaviors from the reward signal alone. Recent works address this by introducing auxiliary representation alignment mechanisms within single-stage reinforcement learning frameworks \cite{wang2024cts, song2026gait}.

\subsection{Fault-Tolerant Locomotion}
\label{subsec:ftloco}

Traditional model-based approaches to fault-tolerant locomotion typically rely on tailored models for specific failure cases and extensive parameter tuning \cite{cui2022fault, chen2022fault}. This limited scalability has motivated the exploration of deep \ac{RL} as a more flexible alternative. Early learning-based approaches assumed that the robot has access to explicit fault state information, enabling policies to condition their behavior on the known damage configuration \cite{yang2021faultaware}. However, this assumption is rarely realistic in practice, where faults must be managed without ground-truth knowledge of their nature or severity. The first works addressing fault-unaware locomotion resorted to model-based meta-\ac{RL} \cite{anne2021meta} and \ac{RL} with adaptive dynamics randomization \cite{okamoto2021acdr}, achieving promising results but remaining limited to simulation deployment.

A prominent direction involves learning to implicitly infer fault conditions by leveraging privileged information~---~such as the ground-truth joint fault status~---~available only in simulation, during training \cite{liu2024towards, luo2023ftnet, kim2024learning, xu2025acl, fu2025contrastive}. These approaches typically rely on teacher-student or asymmetric learning schemes, where privileged signals guide the training of a deployable policy that must rely solely on proprioceptive observation at test time. Within this paradigm, prior works have considered diverse fault types, including joint locking \cite{liu2024towards}, partial or complete power loss \cite{luo2023ftnet}, and even multiple simultaneous faults affecting one or two legs \cite{xu2025acl}. Explored strategies range from latent representation learning with auxiliary regression objectives \cite{luo2023ftnet, kim2024learning, liu2024towards}, to multi-expert distillation \cite{xu2025acl} and contrastive forward prediction learning \cite{fu2025contrastive}. Progressive curriculum learning is also commonly adopted to gradually increase diversity and severity of impairments while mitigating catastrophic forgetting \cite{kim2024learning, lee2025dreamflex}.

Learning a single policy that both identifies the fault condition and rapidly adapts the locomotion pattern is not trivial, as such, several architectures that bypass this issue have been proposed. Some works incorporate an explicit fault estimation module whose predictions are used online to modulate the locomotion policy \cite{lee2025dreamflex}. Others rely on hierarchical or modular strategies, where task-specific policies are independently trained for healthy and faulty conditions and combined through a higher-level selection or coordination mechanisms \cite{hou2024multitask, pei2025ftcpg}.

Despite these advances, most prior works adopt blind locomotion strategy \cite{luo2023ftnet, lee2025dreamflex, fu2025contrastive}, where terrain information is treated as privileged during training and the deployed policy relies solely on proprioceptive feedback. While this avoids integrating perception into the control pipeline, it inherently produces purely reactive behaviors: the robot responds to terrain variations only after physical interaction, rather than planning footholds in advance. Such reactive strategies are typically effective on noisy or mildly uneven terrains, but struggle in more demanding scenarios. This limitation becomes particularly critical for heavier robot platforms operating under hardware faults, where reduced actuation capacity and stability margins leave little tolerance for aggressive corrective actions.

Moreover, several approaches introduce ad-hoc reward terms tailored to specific failure scenarios, such as penalties on ground contact of an impaired leg \cite{lee2025dreamflex, xu2025acl} or additional objectives designed to enforce balance under damage \cite{luo2023ftnet}. Although these terms can stabilize training, they may also bias the learned behavior toward overly constrained or unnatural locomotion patterns. In addition, many works rely on predefined gait generators that encourage constant gait timing \cite{kim2024learning, lee2025dreamflex}. While such priors can produce smooth and natural-looking motion in nominal condition, they restrict adaptability under faults and in uneven terrain, where dynamic modulation of gait frequency and structure is critical for load redistribution and balance.

\acresetall

\section{METHOD}
\label{sec:method}

Our objective is to learn a velocity-commanded control policy that enables an $\numlegs$-legged robot to rapidly adapt to sudden power-loss faults of the actuators while traversing uneven terrain. To achieve this, we employ a deep \ac{RL} method based on an asymmetric actor-critic architecture \cite{pinto2017asymmetric}, trained using \ac{PPO} \cite{schulman2017proximal}.

Our approach leverages asymmetric training to bridge the gap between simulation and reality: the critic has access to explicit fault information, while the actor is trained to approximate a privileged latent representation derived from fault-aware observations using only its proprioceptive history. The resulting policy operates at \ctrlf~Hz, providing position targets for the $\numjoints$ joints and a gait frequency term.

\subsection{Proposed Approach}
\label{subsec:approach}

We formulate fault-tolerant locomotion as a partially observable control problem. While full system state is available during training, the deployed policy only has access to a subset of observations. This asymmetry motivates the use of an asymmetric actor-critic framework with privileged information.

\begin{figure}[t]
    \centering
    \includegraphics[width = \linewidth]{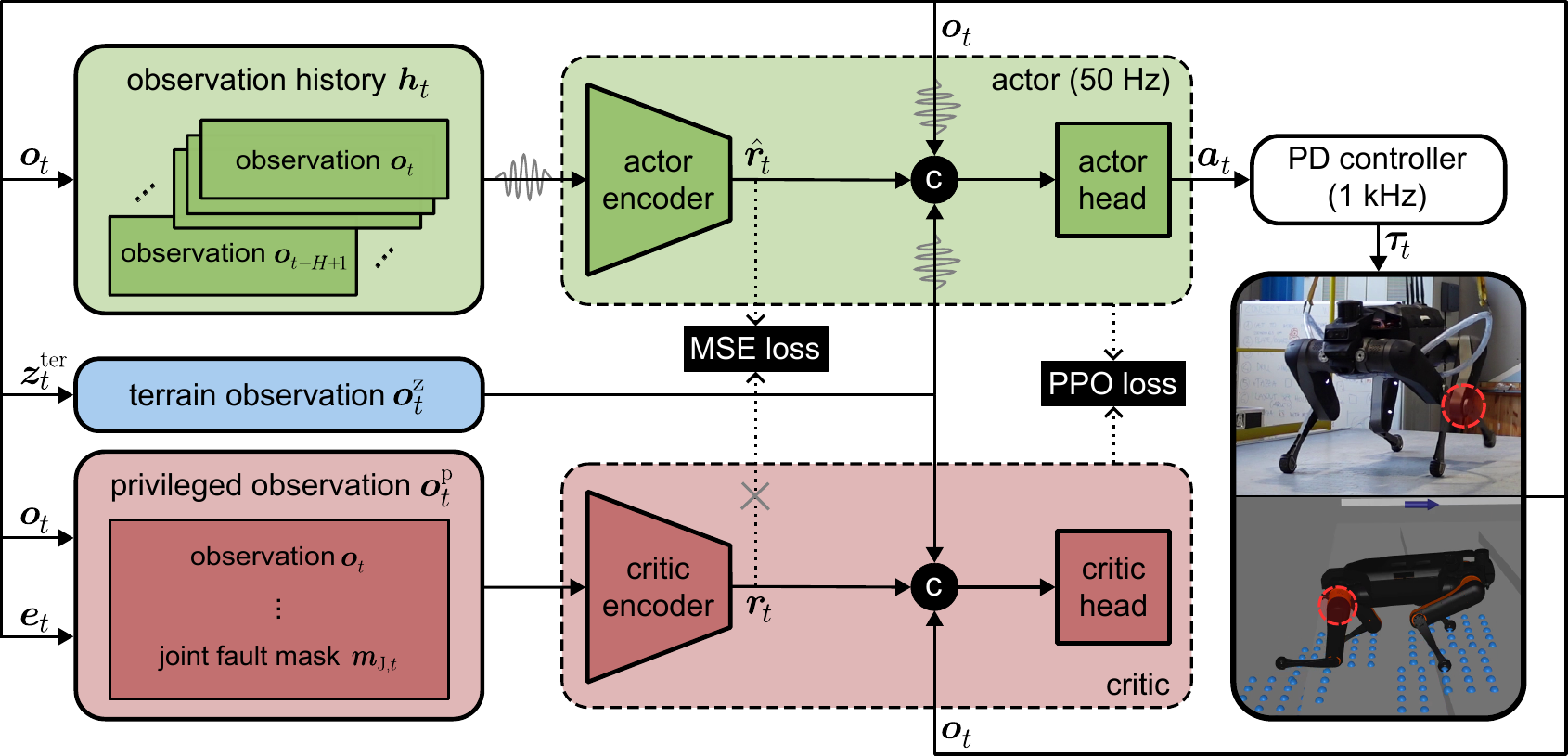}
    \caption{Overview of the proposed training approach based on asymmetric actor-critic framework \cite{pinto2017asymmetric}.
    }
    \label{fig:block_scheme}
\end{figure}

The proposed architecture, illustrated in Fig.~\ref{fig:block_scheme}, consists of an actor and a critic, each comprising an encoder and a head network. The critic encoder embeds the privileged observation $\prvobsat{t}$, composed of proprioceptive observation $\obsat{t}$ and privileged information $\prvinfoat{t}$, into a latent representation $\latentcat{t}$. This latent vector is then concatenated with $\obsat{t}$ and the terrain-related observations $\terrobsat{t}$ to form the input to the critic head for value estimation. In parallel, the actor encoder processes a history of $\histlen$ proprioceptive observations $\obshistat{t}=\tuple{\obsat{t}, \dots, \obsat{t-\histlen+1}}$ and outputs a latent vector $\latentaat{t}$. The actor head receives the concatenation of $\latentaat{t}$, $\obsat{t}$, and $\terrobsat{t}$ to produce the policy action $\actat{t}$.

To bridge the information gap between actor and critic, we introduce an auxiliary latent-alignment objective that encourages $\latentaat{t}$ to approximate $\latentcat{t}$ during training.

\subsection{Action \& Observation  Space}
\label{subsec:obs_act}

The actor head outputs a heterogeneous action vector $\actat{t}=\tuple{\actqat{t}, \actfrequency_t} \in \Reals^{\numjoints+1}$, where $\actqat{t} \in \Reals^{\numjoints}$ represents the joint angle deviations with respect to a default joint configuration $\qdef$, and $\actfrequency_t \in \Reals$ controls the gait frequency.

The reference joint configuration is defined as $\qref_t = \qdef + \actqscale \actqat{t}$, where $\actqscale$ is a scaling factor, and is tracked using a \ac{PD} controller that computes commanded joint torques as $\torque_t=\kpmat(\qref_t-\q_t) - \kdmat\dq_t$, where $\kpmat, \kdmat \in \Reals^{\numjoints \times \numjoints}$ are diagonal gain matrices, and $\q_t, \dq_t \in \Reals^{\numjoints}$ denote the observed joint positions and velocities, respectively.

The scalar action $\actfrequency_t$ modulates the stepping frequency of a reference gait encouraged through a reward term. The reference gait frequency is computed as
\begin{equation*}
    \frequencyref_t=\frequencydef+\frequencyscale \actfrequency_t,
\end{equation*}
where $\frequencydef$ is a nominal frequency and $\frequencyscale$ a scaling coefficient. The reference phase of each leg $\phase_{t,\legid}$ is then updated according to
\begin{equation*}
    \phase_{t+1,\legid}=\operatorname{mod}\bigl(\phase_{t,\legid}+2\pi\ctrltimestep \frequencyref_t+\pi, 2\pi\bigl)-\pi, 
\end{equation*}
with $\ctrltimestep$ denoting the control timestep. The reference phase is initialized according to the selected gait pattern (e.g., trot or walk), reset when the commanded base velocity is near zero, and used to generate a reference foot-contact schedule used in the reward function (see Sect.~\ref{subsec:loss_rew}). Introducing a learnable gait frequency allows the policy to adapt the timing of this reference schedule to terrain geometry and fault conditions, without explicitly redesigning the gait pattern.

\begin{table}[t]
    \caption{Observation Terms. Noise is only added to observations received by the actor.}
    \centering
    \renewcommand{\arraystretch}{1.1}
    \setlength{\extrarowheight}{1pt}
    \setlength{\tabcolsep}{2.2pt}
    \begin{tabularx}{\linewidth}{c|c|X|c|c}
        \specialrule{.05em}{.0em}{.0em}
        \multicolumn{2}{c|}{Term} & Description & Dim & Noise $\unif(\pm \sigma)$ \\
        \specialrule{.1em}{.0em}{.0em}\multirow{8}{*}{$\obsat{t}$} & $\gyro_t$ & base ang. velocity in base frame & $3$ & $0.1$ \\ \cline{2-5}
        & $\grav_t$ & gravity vector in world frame & $3$ & $0.03$ \\ \cline{2-5}
        & $\qdev_t$ & joint angle readings w.r.t. default & $\numjoints$ & $0.05$ \\ \cline{2-5}
        & $\qerr_t$ & joint angle readings w.r.t. reference & $\numjoints$ & $0.05$ \\ \cline{2-5}
        & $\feetposition_t$ & feet position in base frame & $3\numlegs$ & $^\ast$ \\ \cline{2-5}
        & $\actat{t-1}$ & previous action & $\numjoints+1$ & - \\ \cline{2-5}
        & $\cmd_t$ & velocity commands & $3$ & - \\ \cline{2-5}
        & $\phasevec_t$ & reference gait phase & $2\numlegs$ & - \\
        \specialrule{.1em}{.0em}{.0em}\multirow{8}{*}{$\prvinfoat{t}$} & $\lvel_t$ & base lin. velocity in base frame & $3$ & \multirow{8}{*}{-} \\ \cline{2-4}
        & $\lacc_t$ & base lin. acceleration in base frame & $3$ & \\ \cline{2-4}
        & $\avelw_t$ & base ang. velocity in world frame & $3$ & \\ \cline{2-4}
        & $\dq_t$ & joint velocities & $\numjoints$ & \\ \cline{2-4}
        & $\torque_t$ & applied motor torques & $\numjoints$ & \\ \cline{2-4}
        & $\feetcontact_t$ & binary feet contact state & $\numlegs$ & \\ \cline{2-4}
        & $\feetvelocity_t$ & feet lin. velocity & $3\numlegs$ & \\ \cline{2-4}
        & $\jointmaskat{t}$ & binary mask of joint status & $\numjoints$ & \\
        \specialrule{.1em}{.0em}{.0em}\multirow{2}{*}{$\terrobsat{t}$} & $\feetheight_t$ & feet height w.r.t. terrain & $\numlegs$ & $0.005$ \\ \cline{2-5}
        & $\terrainheight_t$ & per-feet terrain height map & $25\numlegs$ & $0.005$ \\
        \specialrule{.05em}{.0em}{.0em}
        \multicolumn{5}{l}{$^\ast$ Noise range for $\feetposition_t$ is $[\pm0.01, \pm0.005, \pm0.02]$ for $x, y, z$ respectively.}
    \end{tabularx}
    \label{tab:observation}
\end{table}

Our architecture utilizes three observation sets: proprioceptive $\obsat{t} \in \Reals^{10+3\numjoints+5\numlegs}$, privileged $\prvobsat{t}=\tuple{\obsat{t}, \prvinfoat{t}} \in \Reals^{19+6\numjoints+9\numlegs}$, and terrain-related $\terrobsat{t} \in \Reals^{26\numlegs}$. All the observed quantities are detailed in Tab.~\ref{tab:observation}.

In addition to sensor-based quantities, the proprioceptive observation $\obsat{t}$ includes: the previous action $\actat{t-1}$, the commands $\cmd_t=\tuple{\xlvcmdat{t},\ylvcmdat{t},\zavcmdat{t}}$~---~i.e., linear velocity in $xy$-plane and angular velocity around $z$-axis~---~, and the encoded gait phase $\phasevec_t=\tuple{\cos{\phase_{t,1}},\sin{\phase_{t,1}},\dots,\cos{\phase_{t,\numlegs}},\sin{\phase_{t,\numlegs}}}$.
The privileged observation $\prvobsat{t}$ augments this set with quantities that are unavailable at deployment, including joint velocities, contact information, and the joint fault mask. The latter encodes the joint status in a binary vector $\jointmaskat{t} \in \squiggly{0,1}^{\numjoints}$, where the $\jointid$-th element is
\begin{equation}
    \maskscalar_{\subjoint,t,\jointid} = 
                    \begin{cases}
                    1,\, \text{if $\jointid$-th joint is healthy}, \\
                    0,\, \text{otherwise}
                    \end{cases}
\end{equation}

Finally, the terrain observation set $\terrobsat{t}$ includes the feet height from the ground $\feetheight_t$ and a per-feet terrain height map $\terrainheight_t$, consisting of 25$\numlegs$ scan points, arranged in $\numlegs$ 5-by-5 grids, centered at the feet $xy$ positions and with 5~cm resolution. Since the focus of this work is adaptation under joint failures on uneven terrains, we assume access to the terrain height information during both training and deployment. In practice, such information can be reconstructed through a perception pipeline processing data from exteroceptive sensors (e.g., LiDAR or RGB-D cameras \cite{miki2022learning, song2026gait}). Integrating a perception module within the proposed framework constitutes a natural extension of this work.

During training, sensor-based observations provided to the actor are corrupted with uniform noise (see Tab.~\ref{tab:observation}), while the critic always receives noiseless quantities.

\subsection{Loss \& Reward Function}
\label{subsec:loss_rew}

The model is trained using the \ac{PPO} objective \cite{schulman2017proximal} augmented with a latent-alignment term. The \ac{PPO} loss is defined as
\begin{equation*}
    \ppoloss=\surrloss + \valw \valloss + \entropyw \entropyloss,
\end{equation*}
where $\surrloss, \valloss, \entropyloss$ denote the surrogate policy loss, value function loss, and entropy regularization term, respectively, and $\valw$, $\entropyw$ are weighting coefficients. To align the actor latent representation with the privileged embedding produced by the critic encoder, we introduce an auxiliary \ac{MSE} loss:
\begin{equation*}
    \mseloss(\latentaat{t},\latentcat{t}) = \Expectation[\squarediff{\latentaat{t}}{\latentcat{t}}]
\end{equation*}
The overall objective is therefore
\begin{equation*}
    \loss = \ppoloss + \msew \mseloss,
\end{equation*}
where $\msew$ controls the strength of the latent-alignment term.

The reward function comprises standard locomotion objectives encouraging velocity tracking, energy efficiency, action smoothness, and safe interaction with the terrain. These terms are designed following the locomotion task rewards of MuJoCo Playground \cite{zakka2025mujocoplayground}.
In addition, we include a phase-consistency reward that promotes agreement between a reference contact schedule and the measured foot contacts. For each leg $\legid$, the reference contact is defined as
\begin{equation}
    \contact_{t,\legid}^\subref = 
                    \begin{cases}
                    0,\, \text{if $\phase_{t,\legid} \in (-\pi,0]$}, \\
                    1,\, \text{if $\phase_{t,\legid} \in (0,\pi]$}
                    \end{cases}
\end{equation}
Faulty legs do not contribute to this reward term. This prevents penalizing unavoidable contact mismatches under failure and allows the policy to autonomously discover effective compensation strategies, without explicitly enforcing a particular faulty-leg behavior.
Finally, a standing posture term is activated only when zero velocity is commanded and only in fully healthy conditions. This avoids interfering with the adaptation mechanism during fault scenarios, while still promoting stable nominal standing behavior. All reward terms are detailed in Tab.~\ref{tab:reward}.
\begin{table}
    \caption{Reward Function Terms}
    \centering
    \begin{tabularx}{\linewidth}{@{} l X c @{}}
        \specialrule{.1em}{.2em}{.2em}
        Term & Formula & Weight \\
        \specialrule{.05em}{.2em}{.2em}$xy$ linear velocity & $\exp \biggl(-\frac{\sqnorm{\xylvcmd-\xylvel}}{0.25}\biggl)$ & $2$ \\
        \specialrule{0em}{.2em}{.2em}$z$ angular velocity & $\exp \biggl(-\frac{(\zavcmd-\zav)^2}{0.25}\biggl)$ & $1.2$ \\
        \specialrule{.05em}{.2em}{.2em}torques & $\norm{\torque} + \sum_\jointid{\abs{\torquei}}$ & $-2\text{e-}4$ \\
        \specialrule{0em}{.2em}{.2em}energy & $\sum_\jointid{\abs{\dqi}\abs{\torquei}}$ & $-1\text{e-}3$ \\
        \specialrule{0em}{.2em}{.2em}action smoothness & $\sqnorm{\actqat{t}-\actqat{t-1}} + \sqnorm{\actqat{t}-2\actqat{t-1}+\actqat{t-2}}$ & $-0.01$ \\
        \specialrule{0em}{.2em}{.2em}termination & $\One_{\squiggly{\text{base/terrain collision}}}$ & $-1$ \\
        \specialrule{0em}{.2em}{.2em}undesired contacts & $\sum_\legid\One_{\squiggly{\legid\text{-th shank/terrain collision}}}$ & $-0.5$ \\
        \specialrule{0em}{.2em}{.2em}feet slide & $\sum_\legid \footcontact\sqnorm{\footvelocity}$ & $-0.1$ \\
        \specialrule{0.05em}{.2em}{.2em}feet phase$^{\ast}$ & $\exp{\bigl(-\sqnorm{\feetcontact-\One_{\squiggly{\norm{\cmd}>0.1}}\feetcontactref}\bigl)}$ & $0.5$ \\
        \specialrule{0em}{.2em}{.2em}standing posture$^{\ast\ast}$ & $\exp\bigl(-10 \cdot \One_{\squiggly{\norm{\cmd}<0.01}}\sqnorm{\q-\qdef}\bigl)$ & $0.1$ \\
        \specialrule{.1em}{.2em}{.2em}
        \multicolumn{3}{l}{$^{\ast}$ contact error of faulty legs is not considered.} \\
        \multicolumn{3}{l}{$^{\ast\ast}$ active in fully healthy scenarios only.}
    \end{tabularx}
    \label{tab:reward}
\end{table}

\subsection{Fault Application Logic}
\label{subsec:fault}

Weakened motor faults are modeled by scaling the \ac{PD} controller output torque $\torque$ using a torque efficiency vector $\efficiency_{\torque} \in [0,1]^{\numjoints}$, both during training and deployment. During training, for each environment and episode of duration $\episodelen$, we uniformly sample a time instant $\tfault \sim \ucal(0,\episodelen)$ and a joint $\jointid \sim \ucal(1,\dots,\numjoints)$ representing the fault onset time and affected joint. From $\tfault$ until the end of the episode, the commanded torque of the selected joint is scaled as $\torque_{\jointid} \leftarrow \efficiencyscalar_{\torque,\jointid} \, \torque_{\jointid}$.

Since complete motor loss significantly affects locomotion, we adopt a curriculum strategy to progressively increase fault severity based on tracking performance \cite{lee2025dreamflex}. At the beginning of training, all joints are assigned an initial efficiency $\efficiencyscalar_{\torque,i} = \efficiencyinit > 0$, ensuring that the policy first experiences partial failures. If the episodic averages of the velocity tracking rewards exceed predefined thresholds $\lvrewthreshold$ and $\avrewthreshold$ under the faulty joint condition, and no termination occurs, the fault severity for that joint is increased in the subsequent episode according to
$\efficiencyscalar_{\torque,\jointid} \leftarrow \max\!\left(\efficiencyscalar_{\torque,\jointid} - \efficiencyvar,\, 0\right)$.

\section{EXPERIMENTAL RESULTS}
\label{sec:results}

\subsection{Training Setup}
\label{subsec:training_setup}

\begin{figure}[t]
    \centering
    \includegraphics[width = \linewidth]{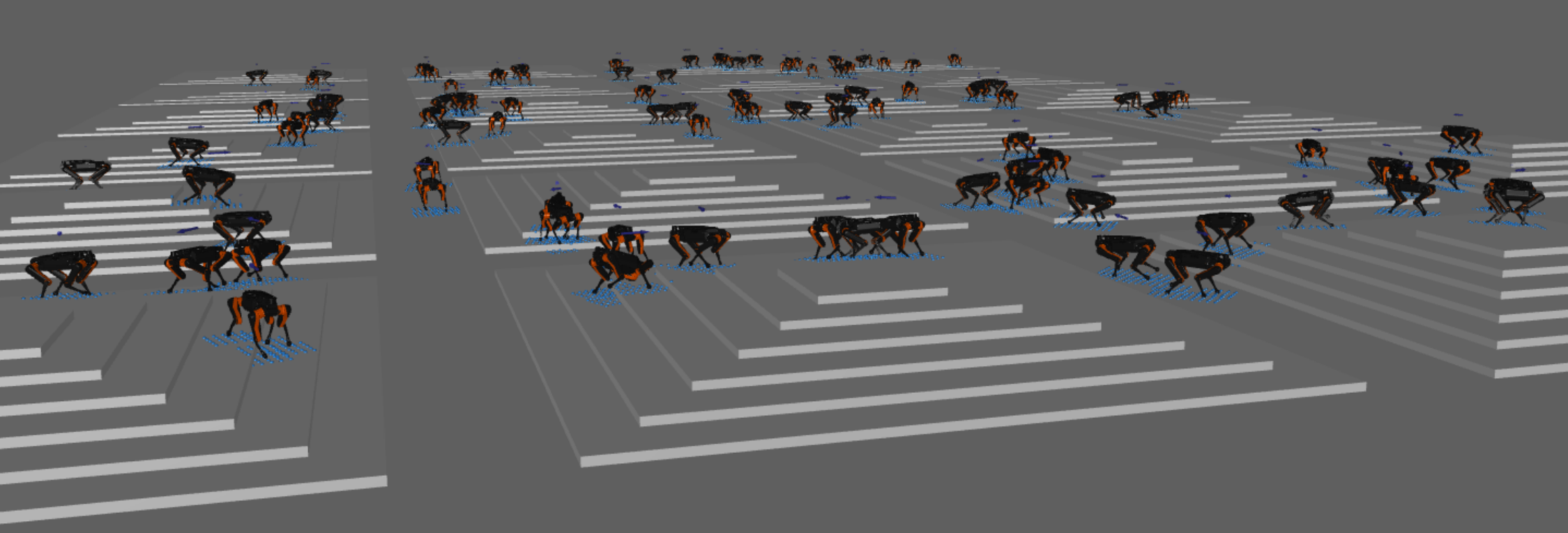}
    \caption{Hundreds of quadrupeds locomoting across the terrain used for training.}
    \label{fig:training_world}
\end{figure}

All trainings were conducted in simulation using MuJoCo XLA (MJX) with the MJWarp physics engine on a NVIDIA GeForce RTX 5090 GPU. We employ a simulation model of the Kyon robot, a 68\,kg quadruped with $\numlegs=4$ legs and $3$ actuated degrees of freedom per leg, resulting in $\numjoints=12$ joints. A total of $8192$ parallel agents are distributed over square stepped pyramids with $10$-by-$10$ meters base, separated by $2$\,m wide flat corridors. Three types of pyramids are considered, with step heights of $4$, $8$, and $12$\,cm, and corresponding widths of $1.2$, $1$, and $0.8$\,m, as shown in Fig.~\ref{fig:training_world}. The maximum episode duration is $\episodelen=20$\,s, corresponding to $1000$ control steps at \ctrlf\,Hz. During each episode, velocity commands $\cmd=\tuple{\xlvcmd,\ylvcmd,\zavcmd}$ are resampled every $4$\,s within the ranges $\xlvcmd \in [-1.5,1.5]$\,m/s, $\ylvcmd \in [-0.8,0.8]$\,m/s, and $\zavcmd \in [-1,1]$\,rad/s. The \ac{PD} gains are set to $\kpmat=300\eye_{12}$ and $\kdmat=10\eye_{12}$. Action scaling parameters are $\actqscale=0.5$, $\frequencyscale=1.25$, with default gait frequency $\frequencydef=1.25$\,Hz. All networks are implemented as MLPs (see Tab.~\ref{tab:networks}), with latent representations $\latentaat{t}$ and $\latentcat{t}$ of dimension $32$. Joint torque efficiency is initially set to $\efficiencyinit=0.25$ and progressively reduced of $\efficiencyvar=0.0125$, as described in Sect.~\ref{subsec:fault}, considering tracking thresholds $\lvrewthreshold=0.7$ and $\avrewthreshold=0.8$. Tab.~\ref{tab:training_params} reports the hyperparameters of the Brax \ac{PPO} implementation~\cite{freeman2021brax} used for training, while Tab.~\ref{tab:domain_rand} summarizes the applied domain randomization.

\begin{table}[t]
\centering
\begin{minipage}{0.48\linewidth}
    \centering
    \caption{Network Sizes}
    \begin{tabular}{lc}
        \toprule
        Module & Hidden Layers \\
        \midrule
        Critic Encoder & $[128, 64]$ \\
        Critic Head & $[256\times5]$ \\
        Actor Encoder & $[512, 128, 64]$ \\
        Actor Head & $[128\times4]$ \\
        \bottomrule
    \end{tabular}
    \label{tab:networks}
\end{minipage}
\hfill
\begin{minipage}{0.48\linewidth}
    \centering
    \caption{PPO Training Settings}
    \begin{tabular}{lc}
        \toprule
        Parameter & Value \\
        \midrule
        Epochs & $4$ \\
        Mini-batches & $32$ \\
        Mini-batch size & $256 \times 20$ \\
        Discount factor $\gamma$ & $0.97$ \\
        GAE factor $\lambda$ & $0.95$ \\
        Clip range & $0.3$ \\
        Learning rate & $3\!\times\!10^{-4}$ \\
        Entropy coeff. $\entropyw$ & $0.01$ \\
        Value coeff. $\valw$ & $0.25$ \\
        \bottomrule
    \end{tabular}
    \label{tab:training_params}
\end{minipage}
\end{table}

\begin{table}
    \caption{Domain Randomization}
    \centering
    \begin{tabular}{lc}
        \toprule
        Term (Unit) & Range \\
        \midrule
        Joint actuator stiffness $\kp$ & $\times U(0.8,\,1.2)$ \\
        Joint actuator damping $\kd$ & $\times U(\sqrt{0.8},\,\sqrt{1.2})$ \\
        Joint passive damping & $\times U(0.8,\,1.2)$ \\
        Joint dry friction & $\times U(0.9,\,1.1)$ \\
        Joint rotor inertia & $\times U(1.0,\,1.05)$ \\
        Terrain sliding friction & $U(0.4,\,1.0)$ \\
        Link masses & $\times U(0.8,\,1.2)$ \\
        Torso mass (kg) & $+U(-5.0,\,10.0)$ \\
        Torso CoM position (m) & $+\,U(-0.10,\,0.10)$ \\
        Initial joint position (rad) & $+\,U(-0.05,\,0.05)$ \\
        \bottomrule
    \end{tabular}
    \label{tab:domain_rand}
\end{table}

\subsection{Performance Evaluation}
\label{subsec:evaluation}

\begin{figure}[t]
    \centering
    \includegraphics[width = \linewidth]{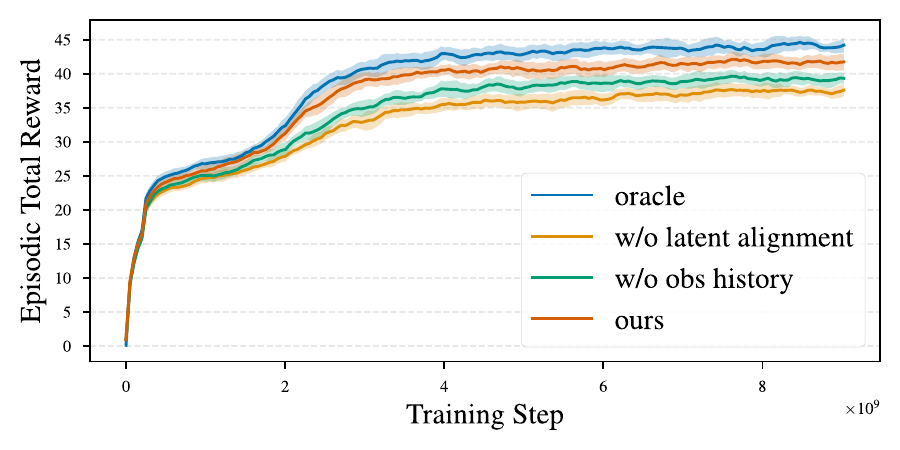}
    \caption{Episodic total reward averaged over five training runs. Shaded regions indicate the 95\% confidence interval. Except for the oracle (blue), which has access to privileged information, our method (red) outperforms the same architecture without proprioceptive history (green) and without latent-alignment loss (yellow).}
    \label{fig:sim_rews}
\end{figure}

We evaluate the proposed method by comparing the following variants:
\begin{itemize}
    \item \textit{Oracle}: both actor and critic have full access to privileged noiseless information, including the joint fault mask, and noiseless terrain observation;
    \item \textit{Ours w/o latent alignment}: the latent-alignment term is disabled by setting $\msew=0$, such that the objective reduces to $\ppoloss$. An observation history length of $\histlen=3$ is used;
    \item \textit{Ours w/o observation history}: the observation history length is set to $\histlen=1$, such that $\obshistat{t}=\obsat{t}$. Latent alignment is enabled by setting $\msew=1$;
    \item \textit{Ours}: the full proposed method with $\histlen=3$ and $\msew=1$.
\end{itemize}
For a fair comparison, we performed five independent training runs for each variant. Fig.~\ref{fig:sim_rews} reports the average episodic total reward during training. The curves indicate that the proposed approach, combining latent representation alignment with proprioceptive observation history, achieves performance closest to the oracle among the compared variants.
We further evaluate the learned policies through simulation experiments conducted on a Laptop equipped with an Intel Core i7-13650HX CPU (2.6\,GHz), with 16\,GB RAM, and an NVIDIA GeForce RTX 4060 GPU. For each training run, we select the policy achieving the highest episodic total reward. Each selected policy is then tested using $1024$ agents randomly distributed across the same terrain used during training (see Fig.~\ref{fig:training_world}) and assigned random velocity commands within the training range. Each experiment lasts $25$\,s and every agent experiences a power-loss fault on a randomly sampled joint at $t=5$\,s.

\begin{figure}[t]
    \centering
    \includegraphics[width = \linewidth]{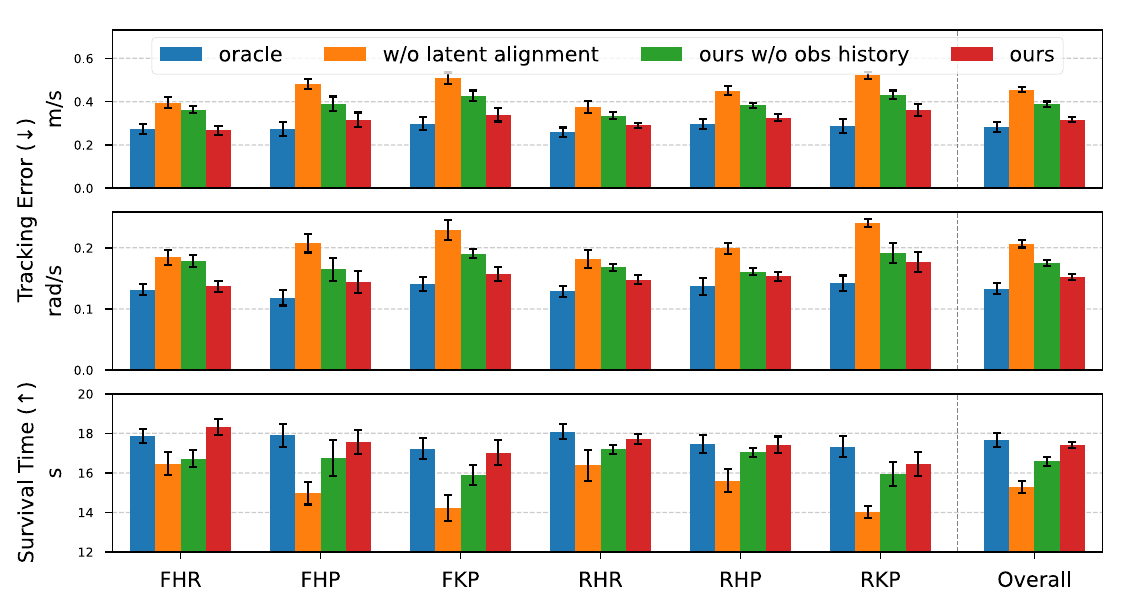}
    \caption{Simulation evaluation under random actuator power loss. Command tracking error (linear and angular) and survival time under faulty conditions (maximum time is $20$\,s) are reported for each method. Results are grouped by fault location (front/rear legs and joint type) and aggregated across all faults. }
    \label{fig:sim_histograms}
\end{figure}

Performance is assessed in terms of linear and angular command tracking errors under fault (lower is better) and survival time under fault (higher is better). Results are summarized in Fig.~\ref{fig:sim_histograms}, where metrics are reported both per fault location~---~distinguishing front and rear legs as well as joint type (hip roll, hip pitch, knee pitch)~---~ and aggregated to provide overall comparison. The results are consistent with the learning curves in Fig.~\ref{fig:sim_rews}, and indicate that the knee-joint faults are generally the most challenging to handle.

\subsubsection{Sim-to-Sim}
\label{subsubsec:simtosim}

\begin{figure}[t]
    \centering
    \includegraphics[width= \linewidth]{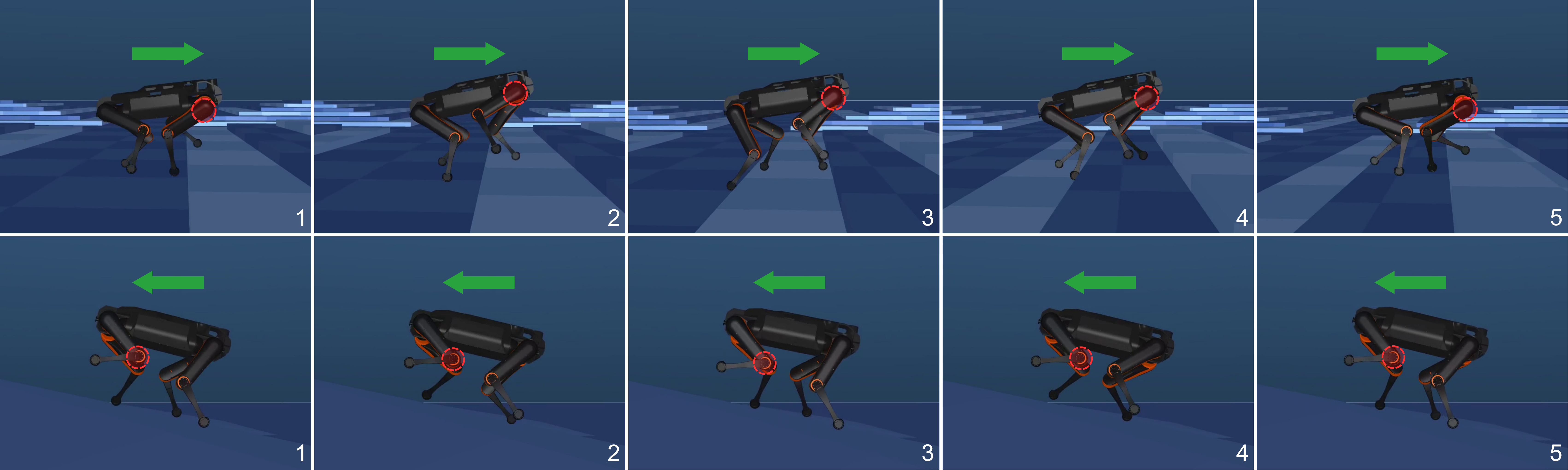}
    \caption{Snapshots from experiments in the MuJoCo simulator with XBot2 integration. Top row: stair climbing on novel step dimensions under power-loss fault at the front-right hip pitch joint. Bottom row: ramp traversal ($13^\circ$) under rear-right knee pitch fault.}
    \label{fig:sim_xbot}
\end{figure}

We evaluated sim-to-sim transferability of the learned policy by considering the MuJoCo simulator \cite{todorov2012mujoco}.
In addition to the simulator loop, low-level communication with each actuator is managed through the XBot2 middleware \cite{laurenzi2023xbot2}. This configuration allows us to replicate the real experimental setup in simulation as closely as possible, aside from unavoidable effects such as model mismatch and unmodeled actuator dynamics.
Specifically, XBot2 operates in a separate process at a custom frequency of 1\,kHz, thereby introducing asynchronous communication between the control policy and the simulator~---~an aspect that was not explicitly accounted for during training.
With this architecture, we aim to reproduce a more realistic deployment scenario, ultimately facilitating a smoother sim-to-real transition.
The policy showed remarkable generalization capabilities even in scenarios not encountered during training, such as stairs with a novel profile~---~e.g., $10$\,cm step height and $0.7$\,m step width~---~and ramps with gradients up to $13^\circ$, as illustrated in Fig.~\ref{fig:sim_xbot}. The top row shows snapshots of the robot climbing stairs following a forward velocity command and experiencing a power-loss fault at the front-right hip pitch joint. The bottom row shows the robot traversing ramps while moving backward and subject to a fault at the rear-right knee joint. Interestingly, faults at the knee often cause the robot to switch to a tripodal locomotion pattern (bottom row), whereas in case of hip-related faults (top row), the injured leg can still contribute to locomotion by exploiting the remaining degrees of freedom to maintain balance and stability.

\subsubsection{Sim-to-Real}
\label{subsubsec:simtoreal}

\begin{figure*}[t]
    \centering
    \includegraphics[width=\linewidth]{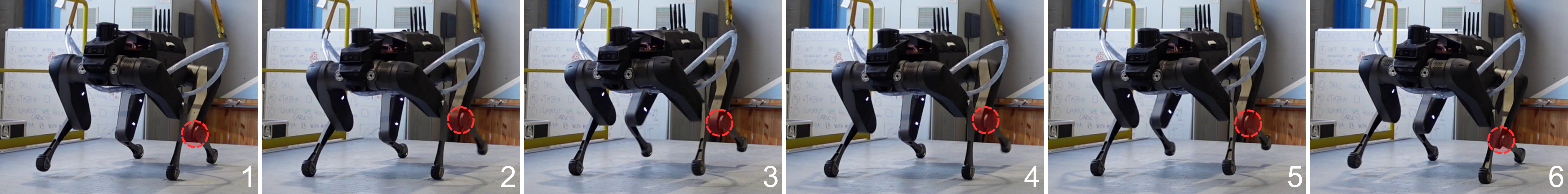}
    \caption{Real-world experimental validation of the proposed control policy conducted on the Kyon quadruped robot. The sequence of snapshots shows successful locomotion under power-loss fault at rear-left knee pitch joint.}
    \label{fig:rlkp_exp_snapshots}
\end{figure*}

Real-world experiments were conducted to further validate our method. Without a perception module, we demonstrate zero-shot sim-to-real transfer on flat terrain by setting the terrain observation $\terrobsat{t}$ provided to the policy to zero.
Fig.~\ref{fig:rlkp_exp_snapshots} shows snapshots of the real robot successfully handling a sudden power-loss fault at the rear-left knee pitch joint while locomoting on flat terrain.

\subsection{Ablation Studies}
\label{subsec:ablation}

\subsubsection{Observation History Length}
\label{subsubsec:obs_history}

\begin{figure}
    \centering
    \includegraphics[width=\linewidth]{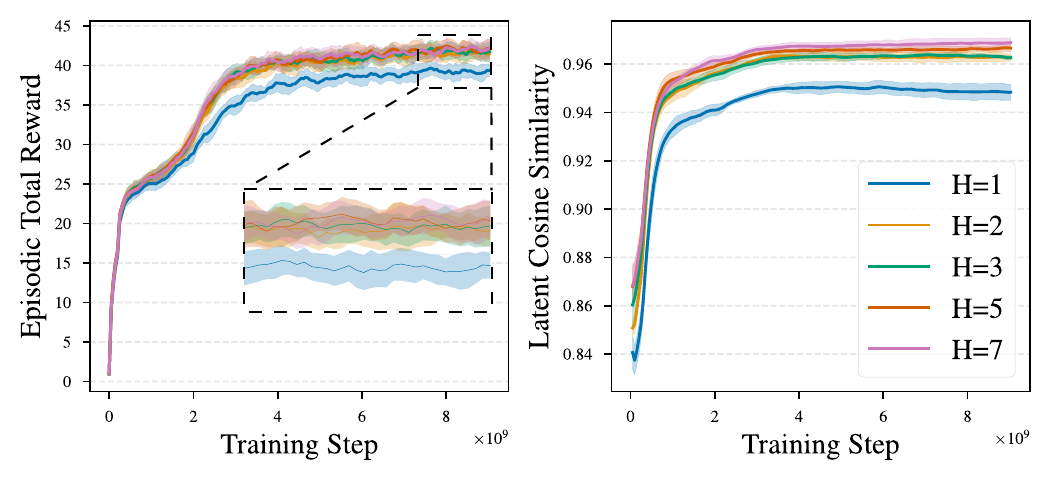}
    \caption{Episodic total reward and latent cosine similarity, averaged over five training runs. Shaded regions denote the 95\% confidence interval. A significant improvement is observed when increasing the proprioceptive history length from $\histlen=1$ to $\histlen=2$, whereas further increases provide marginal or no performance gains.}
    \label{fig:history_comparison}
\end{figure}

We analyze the impact of the proprioceptive observation history by varying the history length $\histlen \in \squiggly{1,2,3,5,7}$. Learning curves are shown in Fig.~\ref{fig:history_comparison}, reporting both the episodic total reward and the cosine similarity between the latent representations $\latentaat{t}$ and $\latentcat{t}$. Introducing past observations helps the policy reconstruct the latent representation of the privileged information. The results indicate that increasing the history length from $\histlen=1$~---~corresponding to no past observations~---~to $\histlen=2$ provides the largest performance improvement. However, further increases in the history length lead to only marginal gains. This suggests that, for the considered task, information about one-step temporal variations is sufficient for the actor to infer the privileged representation. Based on these findings~---~and considering that longer histories increase the number of network parameters~---~we adopt $\histlen=3$ as a practical trade-off.

\subsubsection{Learnable Gait Frequency}
\label{subsubsec:gait_freq}

To assess the benefits of the learnable gait-frequency action, we compare our policy with a free-gait controller that only outputs joint position targets. For a fair comparison, we replace the feet-phase reward term (see Tab.~\ref{tab:reward}) with the widely used feet air-time reward, which encourages natural swing phases and prevents the policy from learning overly conservative low-clearance steps. As in the feet-phase reward, contributions to the feet air-time reward from faulty legs are not considered.

\begin{figure}
    \centering
    \includegraphics[width=\linewidth]{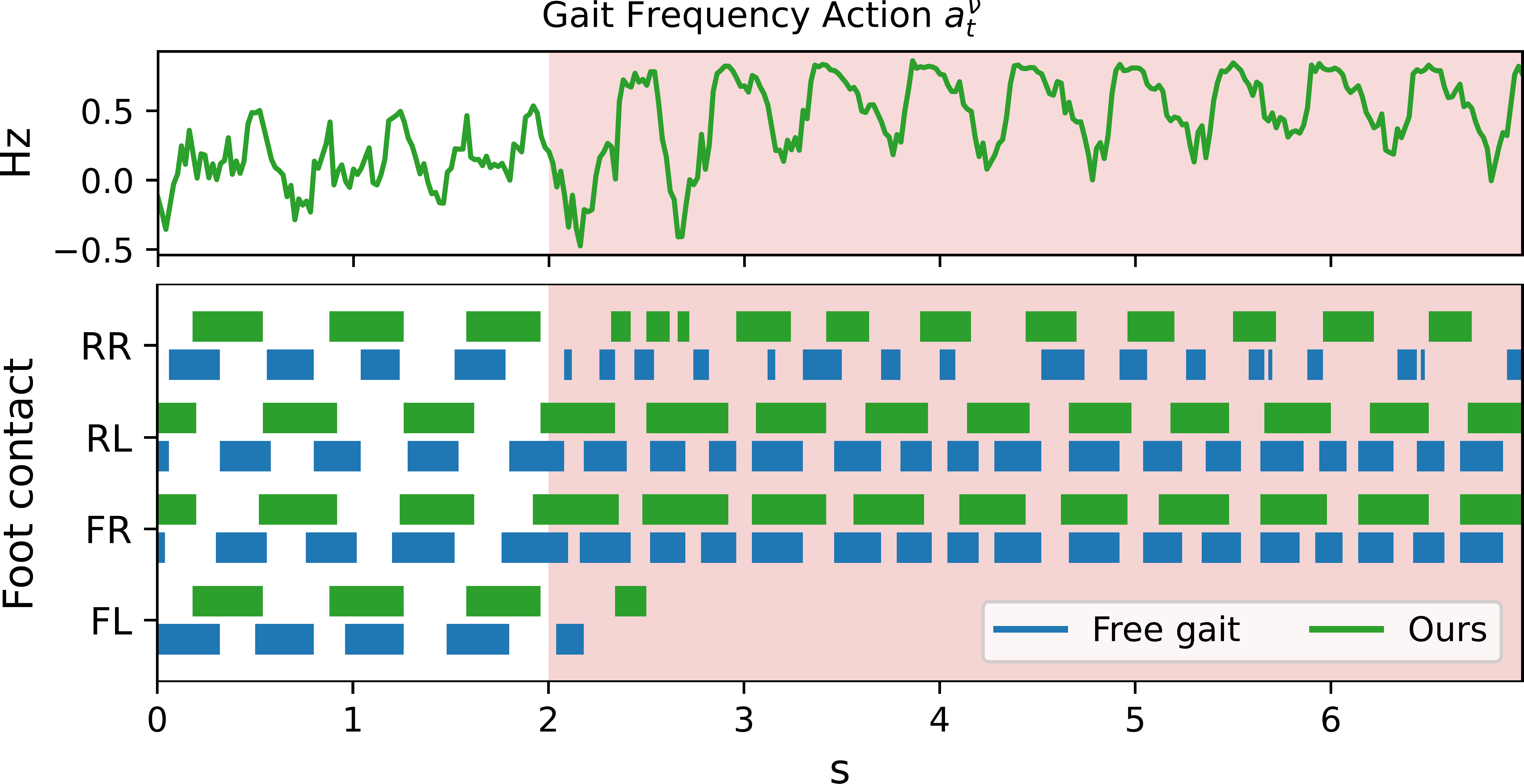}
    \caption{Analysis of the learnable gait frequency. Shaded red regions indicate the faulty condition. Top: profile of the gait-frequency action term $\actfrequency$. Bottom: comparison of the foot-contact sequences generated by our gait-adaptive policy (green) and the free-gait policy (blue). After a fault occurs at the front-left knee joint, the gait-frequency action exhibits a periodic behavior, with higher frequencies during single support and lower frequencies during double support.}
    \label{fig:gait_freq}
\end{figure}

To isolate the effect of fault adaptation, we evaluate the two policies in a flat-terrain scenario where the robot is commanded to move forward at $\xlvcmd=0.7$\,m/s and a fault occurs at the front-left knee pitch joint. Fig.~\ref{fig:gait_freq} (bottom) shows a comparison of the resulting gait patterns. Under faulty conditions, both approaches successfully switch to a tripodal gait. However, while the free-gait policy produces aperiodic and spurious contacts~---~particularly during single support on the rear-right leg~---~our method yields a more stable gait with longer and more periodic swing–stance phases. Moreover, the difference in stance duration between single and double support is clearly reflected in the learned gait-frequency action (top), which increases during single support and decreases during double support.

\begin{figure}

    \centering
    \includegraphics[width=\linewidth]{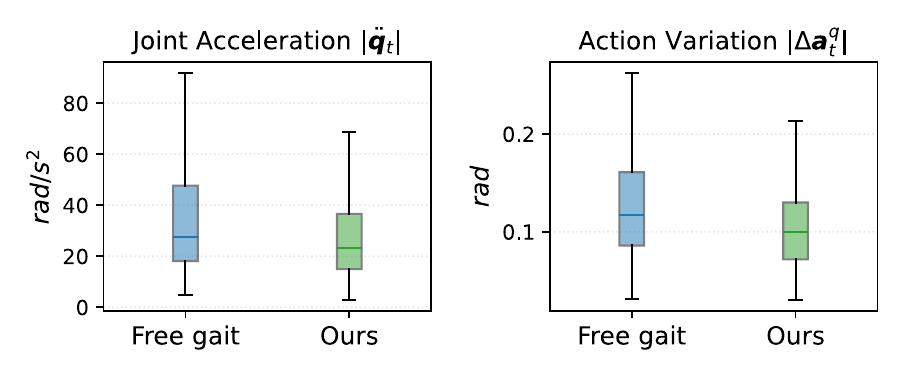}
    \caption{Comparison between our gait-adaptive policy and a free-gait policy. Boxplots of the average joint accelerations (left) and action variation (right), computed over all non-faulty joints. The free-gait policy consistently results in higher values for both metrics, indicating less smooth actuation.}
    \label{fig:boxplots}
\end{figure}

Finally, the smoother and more stable behavior of our controller is further confirmed in Fig.~\ref{fig:boxplots}, where we compare the two approaches in terms of joint accelerations and control action variations, both averaged over the healthy joints during the faulty condition.

\acresetall

\section{CONCLUSIONS}
\label{sec:conclusions}

This work presented a deep reinforcement learning approach for fault-tolerant locomotion under actuator power loss. The method combines an asymmetric actor-critic architecture with a latent representation alignment objective, enabling the actor to infer privileged information from a history of proprioceptive observations.

A central design choice is the inclusion of a learnable gait-frequency action, allowing the policy to adapt step timing. We argue that for heavier quadrupeds, fast reactive strategies effective on lighter platforms do not scale well due to tighter actuation limits and stronger dynamic coupling. Adaptive gait timing therefore becomes critical under degraded actuation. By leveraging terrain information together with gait frequency modulation, the policy achieves robust locomotion without predefined faulty-leg strategies or restrictive reward shaping.

Simulation results validate the importance of proprioceptive history and latent alignment. Sim-to-sim and zero-shot sim-to-real transfer on a 68\,kg quadruped further demonstrate consistent behavior across simulation and hardware. Future work includes integrating onboard perception for terrain reconstruction, learning unified fault-tolerant locomotion and fall-recovery policies, and exploring fault tolerance in hybrid wheeled-legged systems.

\bibliographystyle{IEEEtran}
\bibliography{bibliography}

\end{document}

%% file: tools/acronyms.tex
\acrodef{POMDP}{partially observable Markov decision process}
\acrodef{RL}{Reinforcement Learning}
\acrodef{PPO}{Proximal Policy Optimization}
\acrodef{MSE}{Mean Squared Error}
\acrodef{PD}{proportional-derivative}

%% file: tools/macros.tex
\newcommand{\numjoints}{n_{\subjoint}}
\newcommand{\numlegs}{n_{\subleg}}
\newcommand{\ctrltimestep}{\Delta t}
\newcommand{\jointid}{j}
\newcommand{\legid}{\ell}
\newcommand{\subjoint}{\mathrm{J}}
\newcommand{\subleg}{\mathrm{L}}
\newcommand{\subhgt}{\mathrm{z}}
\newcommand{\subprv}{\mathrm{p}}
\newcommand{\subref}{\mathrm{ref}}
\newcommand{\subdef}{\mathrm{def}}
\newcommand{\subcmd}{\mathrm{cmd}}
\newcommand{\subter}{\mathrm{ter}}
\newcommand{\subfoot}{\mathrm{f}}
\newcommand{\histlen}{H}
\newcommand{\tfault}{\Bar{t}}
\newcommand{\episodelen}{T}
\newcommand{\ctrlf}{50}
\newcommand{\subwrld}{\wcal}

\newcommand{\rew}{r}

\newcommand{\qscalar}{q}                                
\newcommand{\q}{\bm \qscalar}                           
\newcommand{\dqscalar}{\dot \qscalar}                   
\newcommand{\dqi}{\dqscalar_\jointid}                   
\newcommand{\dq}{\dot \q}                               
\newcommand{\scalartorque}{\tau}                        
\newcommand{\torquei}{\scalartorque_\jointid}           
\newcommand{\torque}{\bm \scalartorque}                 
\newcommand{\qref}{\bm{\qscalar}^\subref}               
\newcommand{\qdef}{\bm{\qscalar}^\subdef}               
\newcommand{\qdev}{\Delta\bm{\qscalar}^\subdef}         
\newcommand{\supq}{\text{\qscalar}}                     
\newcommand{\qerr}{\Delta\bm{\qscalar}^\subref}         
\newcommand{\actqat}[1]{\actat{#1}^\supq}               
\newcommand{\actqscale}{s_\supq}                        
\newcommand{\kp}{K_p}
\newcommand{\kd}{K_d}
\newcommand{\kpmat}{\bm{K}_p}                           
\newcommand{\kdmat}{\bm{K}_d}                         

\newcommand{\lspeed}{v}                                 
\newcommand{\aspeed}{\omega}                            
\newcommand{\zav}{\aspeed_z}                            
\newcommand{\cmd}{\bm{v}^\subcmd}                       
\newcommand{\xylvcmd}{\cmd_{xy}}                        
\newcommand{\xlvcmd}{\lspeed_x^\subcmd}
\newcommand{\ylvcmd}{\lspeed_y^\subcmd}
\newcommand{\zavcmd}{\zav^\subcmd}                      
\newcommand{\xlvcmdat}[1]{\lspeed_{#1,x}^\subcmd}       
\newcommand{\ylvcmdat}[1]{\lspeed_{#1,y}^\subcmd}       
\newcommand{\zavcmdat}[1]{\aspeed_{#1,z}^\subcmd}       
\newcommand{\lvel}{\mathbf{v}}                          
\newcommand{\xylvel}{\lvel_{xy}}                        
\newcommand{\lacc}{\dot \lvel}                          
\newcommand{\avel}{\bm{\omega}}                         
\newcommand{\avelw}{^\subwrld\avel}                        

\newcommand{\heightvec}{\bm z}
\newcommand{\terrainheight}{\heightvec^\subter}
\newcommand{\feetheight}{\heightvec^\subfoot}
\newcommand{\feetposition}{\bm{p}^{\subfoot}}           
\newcommand{\footvelocity}{\feetvelocity_\legid}        
\newcommand{\feetvelocity}{\dot{\bm{p}}^{\subfoot}}     
\newcommand{\contact}{c}                                
\newcommand{\footcontact}{c_\legid}                     
\newcommand{\feetcontact}{\bm{c}}                       
\newcommand{\feetcontactref}{\feetcontact^\subref}   

\newcommand{\frequency}{\nu}                            
\newcommand{\actfrequency}{\actscalar^\frequency}       
\newcommand{\frequencyref}{\frequency^\subref}          
\newcommand{\frequencydef}{\frequency^\subdef}          
\newcommand{\frequencyscale}{s_{\frequency}}            
\newcommand{\phase}{\phi}                               
\newcommand{\phasevec}{\bm \Phi}                        

\newcommand{\Value}{V}

\newcommand{\loss}{\lcal}
\newcommand{\ppoloss}{\lcal^\text{PPO}}
\newcommand{\surrloss}{\lcal^\pi}
\newcommand{\valloss}{\lcal^{\Value}}
\newcommand{\valw}{\lambda_1}
\newcommand{\entropyloss}{\lcal^\epsilon}
\newcommand{\entropyw}{\lambda_2}
\newcommand{\mseloss}{\lcal^\text{MSE}}
\newcommand{\msew}{\lambda_3}

\newcommand{\latentc}{\bm r}
\newcommand{\latenta}{\hat{\bm r}}

\newcommand{\obs}{\bm o}
\newcommand{\obshist}{\bm h}
\newcommand{\prvinfo}{\bm e}

\newcommand{\actscalar}{a}
\newcommand{\act}{\bm \actscalar}
\newcommand{\obsat}[1]{\at{\obs}{#1}}                   
\newcommand{\obshistat}[1]{\at{\obshist}{#1}}           
\newcommand{\prvinfoat}[1]{\at{\prvinfo}{#1}}           
\newcommand{\prvobsat}[1]{\obs^{\subprv}_{#1}}          
\newcommand{\terrobsat}[1]{\obs^{\subhgt}_{#1}}        
\newcommand{\actat}[1]{\at{\act}{#1}}                   
\newcommand{\latentcat}[1]{\at{\latentc}{#1}}           
\newcommand{\latentaat}[1]{\at{\latenta}{#1}}           
\newcommand{\gyro}{\bm \omega}
\newcommand{\grav}{\bm g}

\newcommand{\maskscalar}{m}                                     
\newcommand{\mask}{\bm{\maskscalar}}                            
\newcommand{\jointmaskat}[1]{\mask_{\mathrm{J},{#1}}}           
\newcommand{\efficiencyscalar}{k}                               
\newcommand{\efficiency}{\bm{\efficiencyscalar}}                
\newcommand{\efficiencyinit}{\efficiencyscalar^{\text{init}}}   
\newcommand{\efficiencyvar}{\delta_{\efficiencyscalar}}         
\newcommand{\lvrewthreshold}{\rew_{\xylvel}^{\text{th}}}        
\newcommand{\avrewthreshold}{\rew_{\zav}^{\text{th}}}        

\newcommand{\lcal}{\mathcal{L}}

\newcommand{\ucal}{\mathcal{U}}

\newcommand{\wcal}{\mathcal{W}}

\newcommand{\Reals}{\mathbb{R}}

\newcommand{\One}{\mathds{1}}

\newcommand{\tuple}[1]{\langle#1\rangle}
\newcommand{\squiggly}[1]{\left\{#1\right\}}
\newcommand{\norm}[1]{\|#1\|}
\newcommand{\sqnorm}[1]{\|#1\|^2}
\newcommand{\squarediff}[2]{(#1-#2)^2}
\newcommand{\abs}[1]{\left\vert#1\right\vert}
\newcommand{\Expectation}{\mathbb{E}}
\newcommand{\at}[2]{{#1}_{\scriptstyle #2}}     
\newcommand{\unif}{\ucal}
\newcommand{\eye}{\bm I}

%% file: bibliography.bib
@article{lee2020learning,
    author = {Joonho Lee  and Jemin Hwangbo  and Lorenz Wellhausen  and Vladlen Koltun  and Marco Hutter },
    title = {Learning quadrupedal locomotion over challenging terrain},
    journal = {Science Robotics},
    volume = {5},
    number = {47},
    pages = {eabc5986},
    year = {2020},
    doi = {10.1126/scirobotics.abc5986},
    URL = {https://www.science.org/doi/abs/10.1126/scirobotics.abc5986},
    eprint = {https://www.science.org/doi/pdf/10.1126/scirobotics.abc5986}}

@INPROCEEDINGS{yang2021faultaware,
  author={Yang, Fan and Yang, Chao and Guo, Di and Liu, Huaping and Sun, Fuchun},
  booktitle={2021 IEEE 11th Annual International Conference on CYBER Technology in Automation, Control, and Intelligent Systems (CYBER)}, 
  title={Fault-Aware Robust Control via Adversarial Reinforcement Learning}, 
  year={2021},
  volume={},
  number={},
  pages={109-115},
  doi={10.1109/CYBER53097.2021.9588329}}

@misc{okamoto2021acdr,
      title={Reinforcement Learning with Adaptive Curriculum Dynamics Randomization for Fault-Tolerant Robot Control}, 
      author={Wataru Okamoto and Hiroshi Kera and Kazuhiko Kawamoto},
      year={2021},
      eprint={2111.10005},
      archivePrefix={arXiv},
      primaryClass={cs.RO},
      url={https://arxiv.org/abs/2111.10005}, 
}

@misc{pinto2017asymmetric,
      title={Asymmetric Actor Critic for Image-Based Robot Learning}, 
      author={Lerrel Pinto and Marcin Andrychowicz and Peter Welinder and Wojciech Zaremba and Pieter Abbeel},
      year={2017},
      eprint={1710.06542},
      archivePrefix={arXiv},
      primaryClass={cs.RO},
      url={https://arxiv.org/abs/1710.06542}, 
}

@INPROCEEDINGS{anne2021meta,
  author={Anne, Timothée and Wilkinson, Jack and Li, Zhibin},
  booktitle={2021 IEEE/RSJ International Conference on Intelligent Robots and Systems (IROS)}, 
  title={Meta-Learning for Fast Adaptive Locomotion with Uncertainties in Environments and Robot Dynamics}, 
  year={2021},
  volume={},
  number={},
  pages={4568-4575},
  doi={10.1109/IROS51168.2021.9635840}}

@INPROCEEDINGS{liu2024towards,
  author={Liu, Dikai and Yin, Jianxiong and See, Simon},
  booktitle={2024 IEEE Conference on Artificial Intelligence (CAI)}, 
  title={Towards Fault-tolerant Quadruped Locomotion with Reinforcement Learning}, 
  year={2024},
  volume={},
  number={},
  pages={1438-1441},
  doi={10.1109/CAI59869.2024.00257}}

@ARTICLE{luo2023ftnet,
  author={Luo, Zeren and Xiao, Erdong and Lu, Peng},
  journal={IEEE Robotics and Automation Letters}, 
  title={FT-Net: Learning Failure Recovery and Fault-Tolerant Locomotion for Quadruped Robots}, 
  year={2023},
  volume={8},
  number={12},
  pages={8414-8421},
  doi={10.1109/LRA.2023.3329766}}

@INPROCEEDINGS{hou2024multitask,
  author={Hou, Taixian and Tu, Jiaxin and Gao, Xiaofei and Dong, Zhiyan and Zhai, Peng and Zhang, Lihua},
  booktitle={2024 IEEE International Conference on Robotics and Automation (ICRA)}, 
  title={Multi-Task Learning of Active Fault-Tolerant Controller for Leg Failures in Quadruped robots}, 
  year={2024},
  volume={},
  number={},
  pages={9758-9764},
  doi={10.1109/ICRA57147.2024.10610151}}

@misc{schulman2017proximal,
      title={Proximal Policy Optimization Algorithms}, 
      author={John Schulman and Filip Wolski and Prafulla Dhariwal and Alec Radford and Oleg Klimov},
      year={2017},
      eprint={1707.06347},
      archivePrefix={arXiv},
      primaryClass={cs.LG},
      url={https://arxiv.org/abs/1707.06347}, 
}

@INPROCEEDINGS{kim2024learning,
  author={Kim, Mincheol and Shin, Ukcheol and Kim, Jung-Yup},
  booktitle={2024 IEEE International Conference on Robotics and Automation (ICRA)}, 
  title={Learning Quadrupedal Locomotion with Impaired Joints Using Random Joint Masking}, 
  year={2024},
  volume={},
  number={},
  pages={9751-9757},
  doi={10.1109/ICRA57147.2024.10610088}}

@INPROCEEDINGS{lee2025dreamflex,
  author={Lee, Seunghyun and Nahrendra, I Made Aswin and Lee, Dongkyu and Yu, Byeongho and Oh, Minho and Lee, Hyeonwoo and Myung, Hyun},
  booktitle={2025 IEEE International Conference on Robotics and Automation (ICRA)}, 
  title={DreamFLEX: Learning Fault-Aware Quadrupedal Locomotion Controller for Anomaly Situation in Rough Terrains}, 
  year={2025},
  volume={},
  number={},
  pages={16001-16007},
  doi={10.1109/ICRA55743.2025.11127805}}

@ARTICLE{pei2025ftcpg,
  author={Zhang, Pei and Hua, Zhaobo and Qiu, Qiyu and Ding, Jinliang},
  journal={IEEE Robotics and Automation Letters}, 
  title={FT-CPG: Learning Central Pattern Generators for Fault-Tolerant Quadruped Locomotion Under Multi-Joint Failures}, 
  year={2025},
  volume={10},
  number={7},
  pages={6936-6943},
  doi={10.1109/LRA.2025.3572772}}

@article{laurenzi2023xbot2,
title = {The XBot2 real-time middleware for robotics},
journal = {Robotics and Autonomous Systems},
volume = {163},
pages = {104379},
year = {2023},
issn = {0921-8890},
doi = {https://doi.org/10.1016/j.robot.2023.104379},
url = {https://www.sciencedirect.com/science/article/pii/S0921889023000180},
author = {Arturo Laurenzi and Davide Antonucci and Nikos G. Tsagarakis and Luca Muratore}
}

@misc{zakka2025mujocoplayground,
  title = {MuJoCo Playground: An open-source framework for GPU-accelerated robot learning and sim-to-real transfer.},
  author = {Zakka, Kevin and Tabanpour, Baruch and Liao, Qiayuan and Haiderbhai, Mustafa and Holt, Samuel and Luo, Jing Yuan and Allshire, Arthur and Frey, Erik and Sreenath, Koushil and Kahrs, Lueder A. and Sferrazza, Carlo and Tassa, Yuval and Abbeel, Pieter},
  year = {2025},
  publisher = {GitHub},
  url = {https://github.com/google-deepmind/mujoco_playground}
}

@article{cui2022fault,
  title={Fault-tolerant motion planning and generation of quadruped robots synthesised by posture optimization and whole body control},
  author={Cui, Junwen and Li, Zhan and Qiu, Jing and Li, Tianxiao},
  journal={Complex \& Intelligent Systems},
  volume={8},
  number={4},
  pages={2991--3003},
  year={2022},
  publisher={Springer}
}

@INPROCEEDINGS{hutter2016anymal,
  author={Hutter, Marco and Gehring, Christian and Jud, Dominic and Lauber, Andreas and Bellicoso, C. Dario and Tsounis, Vassilios and Hwangbo, Jemin and Bodie, Karen and Fankhauser, Peter and Bloesch, Michael and Diethelm, Remo and Bachmann, Samuel and Melzer, Amir and Hoepflinger, Mark},
  booktitle={2016 IEEE/RSJ International Conference on Intelligent Robots and Systems (IROS)}, 
  title={ANYmal - a highly mobile and dynamic quadrupedal robot}, 
  year={2016},
  volume={},
  number={},
  pages={38-44},
  doi={10.1109/IROS.2016.7758092}}

@inproceedings{todorov2012mujoco,
  title={MuJoCo: A physics engine for model-based control},
  author={Todorov, Emanuel and Erez, Tom and Tassa, Yuval},
  booktitle={2012 IEEE/RSJ International Conference on Intelligent Robots and Systems},
  pages={5026--5033},
  year={2012},
  organization={IEEE},
  doi={10.1109/IROS.2012.6386109}
}

@INPROCEEDINGS{shin2022hound,
  author={Shin, Young-Ha and Hong, Seungwoo and Woo, Sangyoung and Choe, JongHun and Son, Harim and Kim, Gijeong and Kim, Joon-Ha and Lee, KangKyu and Hwangbo, Jemin and Park, Hae-Won},
  booktitle={2022 International Conference on Robotics and Automation (ICRA)}, 
  title={Design of KAIST HOUND, a Quadruped Robot Platform for Fast and Efficient Locomotion with Mixed-Integer Nonlinear Optimization of a Gear Train}, 
  year={2022},
  volume={},
  number={},
  pages={6614-6620},
  doi={10.1109/ICRA46639.2022.9811755}}

@InProceedings{gehring2021inspection,
author="Gehring, C.
and Fankhauser, P.
and Isler, L.
and Diethelm, R.
and Bachmann, S.
and Potz, M.
and Gerstenberg, L.
and Hutter, M.",
editor="Ishigami, Genya
and Yoshida, Kazuya",
title="ANYmal in the Field: Solving Industrial Inspection of an Offshore HVDC Platform with a Quadrupedal Robot",
booktitle="Field and Service Robotics",
year="2021",
publisher="Springer Singapore",
address="Singapore",
pages="247--260",
isbn="978-981-15-9460-1"
}

@ARTICLE{suarez2024eurobin,
  author={Suarez, Alejandro and Kartmann, Rainer and Leidner, Daniel and others},
  journal={IEEE Robotics \& Automation Magazine}, 
  title={Door-to-Door Parcel Delivery From Supply Point to User’s Home With Heterogeneous Robot Team: The euROBIN First-Year Robotics Hackathon}, 
  year={2025},
  volume={32},
  number={3},
  pages={8-25},
  doi={10.1109/MRA.2024.3501954}}

@ARTICLE{hooks2020alphred,
  author={Hooks, Joshua and Ahn, Min Sung and Yu, Jeffrey and Zhang, Xiaoguang and Zhu, Taoyuanmin and Chae, Hosik and Hong, Dennis},
  journal={IEEE Robotics and Automation Letters}, 
  title={ALPHRED: A Multi-Modal Operations Quadruped Robot for Package Delivery Applications}, 
  year={2020},
  volume={5},
  number={4},
  pages={5409-5416},
  doi={10.1109/LRA.2020.3007482}}

@INPROCEEDINGS{solmaz2024rescue,
  author={Solmaz, Selim and Innerwinkler, Pamela and Wójcik, Michał and Tong, Kailin and Politi, Elena and Dimitrakopoulos, George and Purucker, Patrick and Höß, Alfred and Schuller, Björn W. and John, Reiner},
  booktitle={2024 IEEE International Symposium on Robotic and Sensors Environments (ROSE)}, 
  title={Robust Robotic Search and Rescue in Harsh Environments: An Example and Open Challenges}, 
  year={2024},
  volume={},
  number={},
  pages={1-8},
  doi={10.1109/ROSE62198.2024.10591144}}

@ARTICLE{kashiri2019centauro,
  author={Kashiri, Navvab and Baccelliere, Lorenzo and Muratore, Luca and Laurenzi, Arturo and Ren, Zeyu and Hoffman, Enrico Mingo and Kamedula, Malgorzata and Rigano, Giuseppe Francesco and Malzahn, Jorn and Cordasco, Stefano and Guria, Paolo and Margan, Alessio and Tsagarakis, Nikos G.},
  journal={IEEE Robotics and Automation Letters}, 
  title={CENTAURO: A Hybrid Locomotion and High Power Resilient Manipulation Platform}, 
  year={2019},
  volume={4},
  number={2},
  pages={1595-1602},
  doi={10.1109/LRA.2019.2896758}}

@ARTICLE{english1998fault,
  author={English, J.D. and Maciejewski, A.A.},
  journal={IEEE Transactions on Robotics and Automation}, 
  title={Fault tolerance for kinematically redundant manipulators: anticipating free-swinging joint failures}, 
  year={1998},
  volume={14},
  number={4},
  pages={566-575},
  doi={10.1109/70.704223}}

@ARTICLE{yang2002fault,
  author={Jung-Min Yang},
  journal={IEEE Transactions on Systems, Man, and Cybernetics, Part C (Applications and Reviews)}, 
  title={Fault-tolerant gaits of quadruped robots for locked joint failures}, 
  year={2002},
  volume={32},
  number={4},
  pages={507-516},
  doi={10.1109/TSMCC.2002.807274}}

@article{gor2018fault,
author = {Mehul M Gor and PM Pathak and AK Samantaray and Jung Ming Yang and SW Kwak},
title ={Fault-tolerant control of a compliant legged quadruped robot for free swinging failure},
journal = {Proceedings of the Institution of Mechanical Engineers, Part I: Journal of Systems and Control Engineering},
volume = {232},
number = {2},
pages = {161-177},
year = {2018},
doi = {10.1177/0959651817743410},
URL = {https://doi.org/10.1177/0959651817743410},
eprint = {https://doi.org/10.1177/0959651817743410}
}

@article{chen2022fault,
title = {Fault-tolerant gait design for quadruped robots with one locked leg using the GF set theory},
journal = {Mechanism and Machine Theory},
volume = {178},
pages = {105069},
year = {2022},
issn = {0094-114X},
doi = {https://doi.org/10.1016/j.mechmachtheory.2022.105069},
url = {https://www.sciencedirect.com/science/article/pii/S0094114X22003160},
author = {Zhijun Chen and Qingxing Xi and Feng Gao and Yue Zhao}
}

@ARTICLE{valsecchi2023barry,
  author={Valsecchi, Giorgio and Rudin, Nikita and Nachtigall, Lennart and Mayer, Konrad and Tischhauser, Fabian and Hutter, Marco},
  journal={IEEE Robotics and Automation Letters}, 
  title={Barry: A High-Payload and Agile Quadruped Robot}, 
  year={2023},
  volume={8},
  number={11},
  pages={6939-6946},
  doi={10.1109/LRA.2023.3313923}}

@article{miki2022learning,
author = {Takahiro Miki  and Joonho Lee  and Jemin Hwangbo  and Lorenz Wellhausen  and Vladlen Koltun  and Marco Hutter },
title = {Learning robust perceptive locomotion for quadrupedal robots in the wild},
journal = {Science Robotics},
volume = {7},
number = {62},
pages = {eabk2822},
year = {2022},
doi = {10.1126/scirobotics.abk2822},
URL = {https://www.science.org/doi/abs/10.1126/scirobotics.abk2822},
eprint = {https://www.science.org/doi/pdf/10.1126/scirobotics.abk2822}}

@article{margolis2024rapid,
author = {Gabriel B. Margolis and Ge Yang and Kartik Paigwar and Tao Chen and Pulkit Agrawal},
title ={Rapid locomotion via reinforcement learning},
journal = {The International Journal of Robotics Research},
volume = {43},
number = {4},
pages = {572-587},
year = {2024},
doi = {10.1177/02783649231224053},
URL = {https://doi.org/10.1177/02783649231224053},
eprint = {https://doi.org/10.1177/02783649231224053}
}

@article{choi2023learning,
author = {Suyoung Choi  and Gwanghyeon Ji  and Jeongsoo Park  and Hyeongjun Kim  and Juhyeok Mun  and Jeong Hyun Lee  and Jemin Hwangbo },
title = {Learning quadrupedal locomotion on deformable terrain},
journal = {Science Robotics},
volume = {8},
number = {74},
pages = {eade2256},
year = {2023},
doi = {10.1126/scirobotics.ade2256},
URL = {https://www.science.org/doi/abs/10.1126/scirobotics.ade2256},
eprint = {https://www.science.org/doi/pdf/10.1126/scirobotics.ade2256}}

@INPROCEEDINGS{cheng2024extreme,
  author={Cheng, Xuxin and Shi, Kexin and Agarwal, Ananye and Pathak, Deepak},
  booktitle={2024 IEEE International Conference on Robotics and Automation (ICRA)}, 
  title={Extreme Parkour with Legged Robots}, 
  year={2024},
  volume={},
  number={},
  pages={11443-11450},
  doi={10.1109/ICRA57147.2024.10610200}}

@article{hoeller2024anymal,
author = {David Hoeller  and Nikita Rudin  and Dhionis Sako  and Marco Hutter },
title = {ANYmal parkour: Learning agile navigation for quadrupedal robots},
journal = {Science Robotics},
volume = {9},
number = {88},
pages = {eadi7566},
year = {2024},
doi = {10.1126/scirobotics.adi7566},
URL = {https://www.science.org/doi/abs/10.1126/scirobotics.adi7566},
eprint = {https://www.science.org/doi/pdf/10.1126/scirobotics.adi7566}}

@ARTICLE{song2026gait,
  author={Song, Haolin and Zhu, Hongbo and Yu, Tao and Liu, Yan and Yuan, Mingqi and Zhou, Wengang and Chen, Hua and Li, Houqiang},
  journal={IEEE Robotics and Automation Letters}, 
  title={Gait-Adaptive Perceptive Humanoid Locomotion With Real-Time Under-Base Terrain Reconstruction}, 
  year={2026},
  volume={},
  number={},
  pages={1-8},
  doi={10.1109/LRA.2026.3664167}}

@misc{xu2025acl,
      title={AcL: Action Learner for Fault-Tolerant Quadruped Locomotion Control}, 
      author={Tianyu Xu and Yaoyu Cheng and Pinxi Shen and Lin Zhao},
      year={2025},
      eprint={2503.21401},
      archivePrefix={arXiv},
      primaryClass={cs.RO},
      url={https://arxiv.org/abs/2503.21401}, 
}

@InProceedings{fu2025contrastive,
  title = 	 {Contrastive Forward Prediction Reinforcement Learning for Adaptive Fault-Tolerant Legged Robots},
  author =       {Fu, Yangqing and Zhang, Yang and Yang, Qiyue and Yan, Liyun and Cao, Zhanxiang and Gao, Yue},
  booktitle = 	 {Proceedings of The 9th Conference on Robot Learning},
  pages = 	 {3285--3303},
  year = 	 {2025},
  editor = 	 {Lim, Joseph and Song, Shuran and Park, Hae-Won},
  volume = 	 {305},
  series = 	 {Proceedings of Machine Learning Research},
  month = 	 {27--30 Sep},
  publisher =    {PMLR},
  url = 	 {https://proceedings.mlr.press/v305/fu25b.html}
}

@ARTICLE{wang2024cts,
  author={Wang, Hongxi and Luo, Haoxiang and Zhang, Wei and Chen, Hua},
  journal={IEEE Robotics and Automation Letters}, 
  title={CTS: Concurrent Teacher-Student Reinforcement Learning for Legged Locomotion}, 
  year={2024},
  volume={9},
  number={11},
  pages={9191-9198},
  doi={10.1109/LRA.2024.3457379}}

@misc{freeman2021brax,
      title={Brax -- A Differentiable Physics Engine for Large Scale Rigid Body Simulation}, 
      author={C. Daniel Freeman and Erik Frey and Anton Raichuk and Sertan Girgin and Igor Mordatch and Olivier Bachem},
      year={2021},
      eprint={2106.13281},
      archivePrefix={arXiv},
      primaryClass={cs.RO},
      url={https://arxiv.org/abs/2106.13281}, 
}

@misc{rossini2026kyon,
      title={KYON: Semi-Modular Wheel-Legged Quadruped With Agile Bimanual Capability}, 
      author={Luca Rossini and Arturo Laurenzi and Francesco Ruscelli and Yifang Zhang and Giovanbattista Gravina and Lorenzo Baccelliere and Corrado Burchielli and Stefano Cordasco and Nikos Tsagarakis},
      year={2026},
      eprint={2606.30243},
      archivePrefix={arXiv},
      primaryClass={cs.RO},
      url={https://arxiv.org/abs/2606.30243}, 
}
